\documentclass[10pt,twocolumn,letterpaper]{article}
\usepackage{cvpr}              

\usepackage{graphicx}
\usepackage{amsmath}
\usepackage{amssymb}
\usepackage{booktabs}

\usepackage{amsfonts}       
\usepackage{nicefrac}       
\usepackage{microtype}      
\usepackage{amsthm}
\usepackage{bm,bbm}
\usepackage{algorithm}
\usepackage{algorithmic}
\usepackage{threeparttable}
\usepackage{xcolor}

\usepackage[pagebackref,breaklinks,colorlinks]{hyperref}

\usepackage[capitalize]{cleveref}
\crefname{section}{Sec.}{Secs.}
\Crefname{section}{Section}{Sections}
\Crefname{table}{Table}{Tables}
\crefname{table}{Tab.}{Tabs.}

\def\cvprPaperID{2548} 
\def\confName{CVPR}
\def\confYear{2022}

\begin{document}

\title{Multi-level Feature Learning for Contrastive Multi-view Clustering}

\author{
Jie Xu$^{1\dagger}$,
Huayi Tang$^{1\dagger}$,
Yazhou Ren$^{1*}$,
Liang Peng$^1$,
Xiaofeng Zhu$^{1,2}$,
Lifang He$^3$\\
{\small $^1$School of Computer Science and Engineering, University of Electronic Science and Technology of China, Chengdu 611731, China}\\
{\small $^2$Shenzhen Institute for Advanced Study, University of Electronic Science and Technology of China, Shenzhen 518000, China}\\
{\small $^3$Department of Computer Science and Engineering, Lehigh University, PA 18015, USA}\\
{\tt\small jiexuwork@outlook.com,tangh4681@gmail.com,yazhou.ren@uestc.edu.cn}\\
{\tt\small larrypengliang@gmail.com,seanzhuxf@gmail.com,lih319@lehigh.edu}
}

\maketitle

\let\thefootnote\relax\footnotetext{$^{\dagger}$Equal contribution. $^{*}$Corresponding author.}

\begin{abstract}
Multi-view clustering can explore common semantics from multiple views and has attracted increasing attention. However, existing works punish multiple objectives in the same feature space, where they ignore the conflict between learning consistent common semantics and reconstructing inconsistent view-private information. In this paper, we propose a new framework of multi-level feature learning for contrastive multi-view clustering to address the aforementioned issue. Our method learns different levels of features from the raw features, including low-level features, high-level features, and semantic labels/features in a fusion-free manner, so that it can effectively achieve the reconstruction objective and the consistency objectives in different feature spaces. Specifically, the reconstruction objective is conducted on the low-level features. Two consistency objectives based on contrastive learning are conducted on the high-level features and the semantic labels, respectively. They make the high-level features effectively explore the common semantics and the semantic labels achieve the multi-view clustering. As a result, the proposed framework can reduce the adverse influence of view-private information. Extensive experiments on public datasets demonstrate that our method achieves state-of-the-art clustering effectiveness.
\end{abstract}

\section{Introduction}\label{intro}
Multi-view clustering (MVC) is attracting more and more attention in recent years \cite{liu2021one,zhong2021improved,Xu_2021_ICCV,yang2022robust} as multi-view data or multi-modal data can provide common semantics to improve the learning effectiveness \cite{hu2019deep,maninis2022vid2cad,peng2020mra,saqur2020multimodal,asano2020labelling,wei2021universal}. In the literature, existing MVC methods can be roughly divided into two categories, \ie, traditional methods and deep methods.


\begin{figure*}[t]
  \centering
   \includegraphics[width=6.4in]{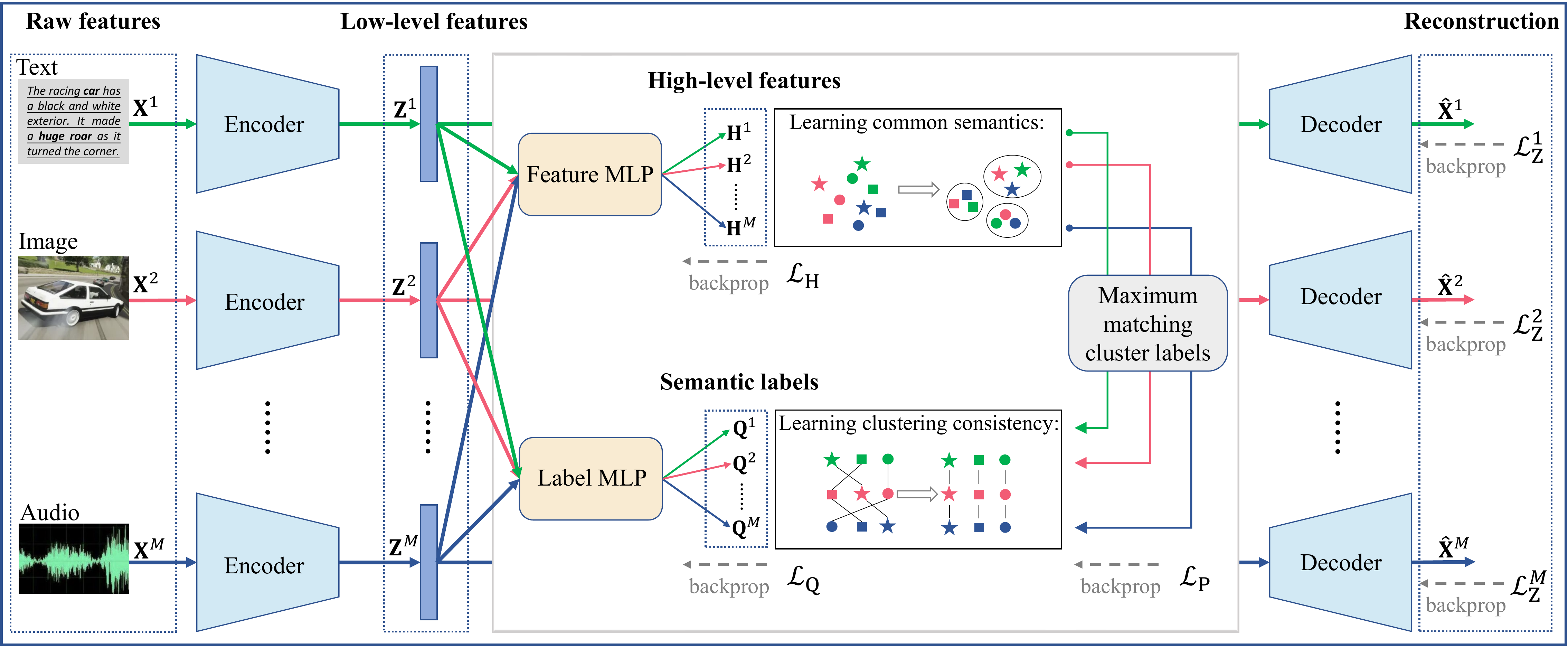}   
   \caption{The framework of MFLVC. We avoid the direct feature fusion in multi-level feature learning, which learns the low-level features $\mathbf Z^m$, the high-level features ${\mathbf H}^m$, and the semantic labels ${\mathbf Q}^m$ from the raw features $\mathbf X^m$ for each view.
   The reconstruction objective $\mathcal{L}_{\mathbf{Z}}^m$ is individually conducted on ${\mathbf Z}^m$.
   Two consistency objectives (\ie, $\mathcal{L}_{\mathbf{H}}$ and $\mathcal{L}_{\mathbf{Q}}$) are conducted on $\{\mathbf{H}^m\}_{m=1}^M$ and $\{\mathbf{Q}^m\}_{m=1}^M$, respectively. 
   Furthermore, $\mathcal{L}_{\mathbf{P}}$ is optimized to leverage the cluster information of $\{\mathbf{H}^m\}_{m=1}^M$ to improve the clustering effectiveness of $\{{\mathbf Q}^m\}_{m=1}^M$.
   }
   \label{fig:framework}
\end{figure*}

The traditional MVC methods conduct the clustering task based on traditional machine learning methods and can be subdivided into three subgroups, including subspace methods \cite{cao2015diversity,luo2018consistent,li2019reciprocal}, matrix factorization methods \cite{zhao2017multi,Wen_2018_ECCV_Workshops,yang2020uniform}, and graph methods \cite{nie2017self,zhan2017graph,zhu2018one}. Many traditional MVC methods have the drawbacks such as poor representation ability and high computation complexity, resulting in limited performance in the complex scenarios with real-world data \cite{guo2019anchors}.

Recently, deep MVC methods have gradually become a popular trend in the community due to the outstanding representation ability \cite{abavisani2018deep,alwassel2019self,li2019deep,yin2020shared,wen2020cdimc,xu2021deep,Xu_2021_ICCV}.
Previous deep MVC methods can be subdivided into two subgroups, \ie, two-stage methods and one-stage methods.
Two-stage deep MVC methods (\eg, \cite{lin2021completer,Xu_2021_ICCV}) focus on separately learning the salient features from multiple views and  performing the clustering task.
However, Xie \etal \cite{xie2016unsupervised} present that the clustering results can be leveraged to improve the quality of feature learning.
Therefore, one-stage deep MVC methods (\eg, \cite{zhou2020end,trostenMVC}) embed the feature learning with the clustering task in a unified framework to achieve end-to-end clustering.

Multi-view data contains two kinds of information, \ie, the common semantics across all views and the view-private information for individual view.
For example, a text and an image can be combined to describe common semantics, while the unrelated context in the text and the background pixels in the image are meaningless view-private information for learning common semantics.
In multi-view learning, it is an always-on topic to learn common semantics and avoid the misleading of meaningless view-private information.
Although important progress has been achieved by existing MVC methods, they have the following drawbacks to be addressed:
(1) Many MVC methods (\eg, \cite{zhou2020end,trostenMVC}) try to discover the latent cluster patterns by fusing the features of all views.
However, the meaningless view-private information might be dominant in the feature fusion process, compared to the common semantics, and thus interferes with the quality of clustering.
(2) Some MVC methods (\eg, \cite{li2019reciprocal,lin2021completer}) leverage the consistency objective on the latent features to explore the common semantics across all views.
However, they usually need the reconstruction objective on the same features to avoid the trivial solution.
This induces the conflict that the consistency objective tries to learn the features with common semantics across all views as much as possible while the reconstruction objective hopes the same features to maintain the view-private information for individual view.

In this paper, we propose a new framework of multi-level feature learning for contrastive multi-view clustering (MFLVC for short) to address the aforementioned issues, as shown in Figure \ref{fig:framework}.
Our goals include (1) designing a fusion-free MVC model to avoid fusing the adverse view-private information among all views and (2) generating different levels of features for the samples in each view including low-level features, high-level features, and semantic labels/features.
To do this, we first leverage the autoencoder to learn the low-level features from raw features, and then obtain the high-level features and semantic labels via stacking two MLPs on low-level features.
Each MLP is shared by all views and is conducive to filtering out the view-private information.
Furthermore, we take the semantic labels as the anchors, which combine with the cluster information in the high-level features to improve the clustering effectiveness.
In this framework, the reconstruction objective is achieved by the low-level features while two consistency objectives are achieved by the high-level features and the semantic labels, respectively.
Moreover, these two consistency objectives are conducted by contrastive learning, which makes the high-level features focus on mining the common semantics across all views and makes the semantic labels represent consistent cluster labels for multi-view clustering, respectively.
As a result, the conflict between the reconstruction objective and two consistency objectives is alleviated.
Compared to previous works, our contributions are listed as follows:

\begin{itemize}
\item
We design a fusion-free MVC method which conducts different objectives in different feature spaces to solve the conflict between the reconstruction and consistency objectives. In this way, our method is able to effectively explore the common semantics across all views and avoid their meaningless view-private information.
\item
We propose a flexible multi-view contrastive learning framework, which can be used to simultaneously achieve the consistency objectives for the high-level features and the semantic labels. The high-level features enjoy good manifolds and represent common semantics, which enable to improve the quality of semantic labels.
\item
Our method is robust to the hyper-parameters' setting due to the well-designed framework. We conduct ablation studies in details, including the loss components and contrastive learning structures to understand the proposed model. Extensive experiments demonstrate that it achieves state-of-the-art clustering effectiveness.
\end{itemize}

\section{Related Work}
\textbf{Multi-view clustering}. The first category of MVC methods belongs to subspace clustering \cite{luo2018consistent,li2019reciprocal}, which focuses on learning a common subspace representation for multiple views.
For instance, the traditional subspace clustering was extended by \cite{cao2015diversity}, where the authors presented a diversity-induced mechanism for multi-view subspace clustering.
The second category of MVC methods is based on the matrix factorization technique \cite{liu2013multi,zhao2017multi} that is formally equivalent to the relaxation of $K$-means~\cite{macqueen1967some}.
For example, Cai \etal. \cite{cai2013multi} introduced a shared clustering indicator matrix for multiple views and handled a constrained matrix factorization problem.
The third category of MVC methods is graph based MVC \cite{nie2017self,peng2019comic}, where graph structures are build to preserve the adjacency relationship among samples.
The fourth category of MVC methods is based on deep learning framework, as known as deep MVC methods, which have been exploited increasingly and can be further roughly divided into two groups, \ie, two-stage deep MVC methods \cite{lin2021completer,Xu_2021_ICCV} and one-stage deep MVC methods \cite{li2019deep,zhou2020end,xu2021self}.
These methods utilize the excellent representation ability of deep neural networks to discover the latent cluster patterns of multi-view data.

\textbf{Contrastive learning}. Contrastive learning \cite{chen2020simple,wang2020understanding} is an attention-getting unsupervised representation learning method, with the idea that maximizing the similarities of positive pairs while minimizing that of negative pairs in a feature space. This learning paradigm has lately achieved promising performance in computer vision, such as \cite{van2020scan,niu2021spice}.
For example, a one-stage online image clustering method was proposed in \cite{li2021contrastive}, which explicitly conducted contrastive learning in the instance-level and cluster-level.
For multi-view learning, there are also some works based on contrastive learning \cite{tian2019contrastive,hassani2020contrastive,lin2021completer,pielawski2020comir}.
For instance, Tian \etal. \cite{tian2019contrastive} proposed a contrastive multi-view coding framework to capture underlying scene semantics.
In \cite{hassani2020contrastive}, the authors developed a multi-view representation learning method to tackle graph classification via contrastive learning.
Recently, some works investigated different contrastive learning frameworks for multi-view clustering \cite{lin2021completer,trostenMVC,pan2021multi}.

\section{Method}
\noindent \textbf{Raw features}. A multi-view dataset $\{\mathbf{X}^m\in \mathbb{R}^{N\times D_m}\}_{m=1}^{M}$ includes $N$ samples across $M$ views, where $\mathbf{x}_i^m \in \mathbb{R}^{D_m}$ denotes the $D_m$-dimensional sample from the $m$-th view. The dataset is treated as the raw features where multiple views have $K$ common cluster patterns to be discovered.

\subsection{Motivation}\label{sec:motivation}
The multi-view data usually have redundancy and random noise, so the mainstream methods always learn salient representations from raw features.
In particular, autoencoder \cite{hinton2006reducing,song2018self} is a widely used unsupervised model and it can project the raw features into a customizable feature space.
Specifically, for the $m$-th view, we denote $E^m{(\mathbf{X}^m;\theta^m)}$ and $D^m{(\mathbf{Z}^m;\phi^m)}$, respectively, as the encoder and the decoder, where $\theta^m$ and $\phi^m$ are network parameters, denote $\mathbf{z}_i^m= E^m{(\mathbf{x}_i^m)} \in \mathbb{R}^{L}$ as the $L$-dimensional latent feature of the $i$-th sample, and denote $\mathcal{L}_{\mathbf{Z}}^m$ as the reconstruction loss between input $\mathbf{{X}}^m$ and output $\mathbf{\hat{X}}^m\in \mathbb{R}^{N\times D_m}$, so the reconstruction objective of all views is formulated as:
\begin{equation}\label{lr}
    \small \mathcal{L}_{\mathbf{Z}} = \sum_{m=1}^M \mathcal{L}_{\mathbf{Z}}^m = \sum_{m=1}^M \sum_{i=1}^N
    \left\|
    {\mathbf{x}_i^m-D^m({E^m{(\mathbf{x}_i^m)})}}
    \right\|
    _2^2.
\end{equation}
Based on $\{\mathbf{Z}^m=E^m(\mathbf{X}^m)\}_{m=1}^M$, MVC aims to mine the common semantics across all views to improve the clustering quality.
To achieve this, existing MVC methods still have two challenges to be addressed:
(1) Many MVC methods (\eg, \cite{li2019deep,zhou2020end}) fuse the features of all views $\{\mathbf{Z}^m\}_{m=1}^M$ to obtain a common representation for all views.
In this way, the multi-view clustering task is transformed to single-view clustering task by conducting clustering directly on the fused features.
However, the features of each view $\mathbf{Z}^m$ contain the common semantics as well as the view-private information.
The latter is meaningless or even misleading, which might interfere with the quality of fused features and result in poor clustering effectiveness.
(2) Some MVC methods (\eg, \cite{lin2021completer,cheng2020multi}) learn consistent multi-view features to explore the common semantics by conducting a consistency objective on $\{\mathbf{Z}^m\}_{m=1}^M$, \eg, minimizing the distance of correlational features across all views.
However, they also apply Eq. (\ref{lr}) to punish constraints on $\{\mathbf{Z}^m\}_{m=1}^M$ to avoid the model collapse and producing trivial solutions \cite{guo2017improved,lin2021completer}.
The consistency objective and the reconstruction objective are pushed on the same features, so that their conflict may limit the quality of $\{\mathbf{Z}^m\}_{m=1}^M$. 
For example, the consistency objective aims to learn the common semantics while the reconstruction objective hopes to maintain the view-private information.

Recently, contrastive learning becomes popular and can be applied to achieve the consistency objective for multiple views.
For instance, Trosten \etal \cite{trostenMVC} proposed a one-stage contrastive MVC method but its feature fusion suffers from challenge (1).
Lin \etal \cite{lin2021completer} presented a two-stage contrastive MVC method by learning consistent features, but it does not consider challenge (2).
Additionally, many contrastive learning methods (\eg, \cite{oord2018representation,van2020scan,li2021contrastive}) mainly handle single-view data with data augmentation. Such specific structure makes it difficult be applied in multi-view scenarios.

To address the aforementioned challenges, we propose a new framework of multi-level feature learning for contrastive multi-view clustering (named MFLVC) as shown in Figure \ref{fig:framework}.
Specially, to reduce the adverse influence of view-private information, our framework avoids the direct feature fusion and builds a multi-level feature learning model for each view.
To alleviate the conflict between the consistency objective and the reconstruction objective, we propose to conduct them in different feature spaces, where the consistency objective is achieved by the following multi-view contrastive learning.



\subsection{Multi-view Contrastive Learning}

Since the features $\{\mathbf{Z}^m\}_{m=1}^M$ obtained by Eq. (\ref{lr}) mix the common semantics with the view-private information,
we treat $\{\mathbf{Z}^m\}_{m=1}^M$ as low-level features and learn another level of features, \ie, high-level features.
To do this, we stack a feature MLP on $\{\mathbf{Z}^m\}_{m=1}^M$ to obtain the high-level features $\{\mathbf{H}^m\}_{m=1}^M$, where $\mathbf{h}_i^m \in \mathbb{R}^{H}$ and the feature MLP is a one-layer linear MLP denoted by $F(\{\mathbf{Z}^m\}_{m=1}^M;\mathbf{W}_H)$.
In the low-level feature space, we leverage the reconstruction objective Eq. (\ref{lr}) to preserve the representation ability of $\{\mathbf{Z}^m\}_{m=1}^M$ so as to avoid the issue of model collapse.
In the high-level feature space, we further achieve the consistency objective by contrastive learning to make $\{\mathbf{H}^m\}_{m=1}^M$ focus on learning the common semantics across all views.

Specifically, each high-level feature $\mathbf{h}_i^m$ has $(MN-1)$ feature pairs, \ie, $\{\mathbf{h}_i^m,\mathbf{h}_j^n\}_{j = 1, \dots, N}^{n = 1, \dots, M}$, where $\{\mathbf{h}_i^m,\mathbf{h}_i^n\}_{n\ne m}$ are $(M-1)$ positive feature pairs and the rest $M(N-1)$ feature pairs are negative feature pairs.
In contrastive learning, the similarities of positive pairs should be maximized and that of negative pairs should be minimized.
Inspired by NT-Xent~\cite{chen2020simple}, the cosine distance is applied to measure the similarity between two features:
\begin{equation}\label{cosine}
    \small d(\mathbf{h}_i^m,\mathbf{h}_j^n)=\frac{\langle \mathbf{h}_i^m,\mathbf{h}_j^n \rangle}{\Vert \mathbf{h}_i^m \Vert \Vert \mathbf{h}_j^n \Vert},
\end{equation}
where $\langle \cdot,\cdot \rangle$ is dot product operator.
Then, the feature contrastive loss between $\mathbf{H}^m$ and $\mathbf{H}^n$ is formulated as:
\begin{equation}\label{fccc}
\small \ell_{fc}^{(mn)}=- \frac{1}{N} \sum_{i=1}^N \log \frac{e^{d(\mathbf{h}_i^m,\mathbf{h}_i^n)/\tau_F}}{\sum_{j=1}^{N}\sum_{v=m,n} {e^{d(\mathbf{h}_i^m,\mathbf{h}_j^v)/\tau_F} - e^{1/\tau_F}}},
\end{equation}
where $\tau_F$ denotes the temperature parameter. In this paper, we design an accumulated multi-view feature contrastive loss across all views as:
\begin{equation}\label{lfr}
    \small \mathcal{L}_{\mathbf{H}}=\frac{1}{2}\sum_{m=1}^M \sum_{n\ne m}\ell_{fc}^{(mn)}.
\end{equation}

In consequence, the features of each view can be written as $\mathbf{H}^m=\mathbf{W}_H \mathbf{Z}^m=\mathbf{W}_H E^m(\mathbf{X}^m)$.
The encoder $E^m$ is conducive to filtering out the random noise of $\mathbf{X}^m$.
The reconstruction objective on $\mathbf{Z}^m$ avoids the model collapse as well as pushes both the common semantics and view-private information to be preserved in $\mathbf{Z}^m$.
$\mathbf{W}_H$ is conducive to filtering out the view-private information of $\{\mathbf{Z}^m\}_{m=1}^M$.
The consistency objective on $\{\mathbf{H}^m\}_{m=1}^M$ allows them to mine the common semantics across all views.
As a result, the clusters of high-level features are close to the true semantic clusters.
Intuitively, semantic information is a high-level concept that does not involve meaningless noise.
Therefore, the high-level features within the same cluster are close to each other, resulting in dense shapes (verified in \cref{sec:Understand}).

\textbf{Learning semantic labels.}
This part explains how to obtain semantic labels for end-to-end clustering from the raw features in a fusion-free model.
Specifically, we obtain the cluster assignments for all views $\{\mathbf{Q}^m\in \mathbb{R}^{N\times K}\}_{m=1}^M$ via a shared label MLP stacked on the low-level features, \ie, $L(\{\mathbf{Z}^m\}_{m=1}^M;\mathbf{W}_Q)$.
The last layer of the label MLP is set to the Softmax operation to output the probability, \eg, $q_{ij}^m$ represents the probability that the $i$-th sample belongs to the $j$-th cluster in the $m$-th view.
Hence, the semantic label is identified by the largest element in a cluster assignment.

In real-world scenarios, however, some views of a sample might have wrong cluster labels due to the misleading of view-private information.
In order to obtain robustness, we need to achieve clustering consistency, \ie, the same cluster labels of all views represent the same semantic cluster.
In other words, $\{\mathbf Q_{\mathbf \cdot j}^m\}_{m=1}^M$ ($\mathbf Q_{\mathbf \cdot j}^m \in \mathbb{R}^{N}$) need to be consistent.
Similar to learning the high-level features, we adopt contrastive learning to achieve this consistency objective.
For the $m$-th view, the same cluster labels $\mathbf Q_{\mathbf \cdot j}^m$ have $(MK-1)$ label pairs, \ie, $\{\mathbf Q_{\mathbf \cdot j}^m,\mathbf Q_{\mathbf \cdot k}^n\}_{k = 1, \dots, K}^{n = 1, \dots, M}$, where $\{\mathbf Q_{\mathbf \cdot j}^m,\mathbf Q_{\mathbf \cdot j}^n\}_{n\ne m}$ are constructed as $(M-1)$ positive label pairs and the rest $M(K-1)$ label pairs are negative label pairs.
We further define the label contrastive loss between $\mathbf Q^m$ and $\mathbf Q^n$ as:
\begin{equation}\label{lccc}
    \small \ell_{lc}^{(mn)}= - \frac{1}{K} \sum_{j=1}^K \log \frac{e^{d(\mathbf{Q}_{\mathbf \cdot j}^m,\mathbf{Q}_{\mathbf \cdot j}^n)/\tau_L}}{\sum_{k=1}^K \sum_{v=m,n} e^{d(\mathbf{Q}_{\mathbf \cdot j}^m,\mathbf{Q}_{\mathbf \cdot k}^v)/\tau_L} - e^{1/{\tau}_L}},
\end{equation}
where $\tau_L$ denotes the temperature parameter.
As thus, the clustering-oriented consistency objective is defined by:
\begin{equation}\label{flc}
    \small \mathcal{L}_{\mathbf{Q}}=\frac{1}{2}\sum_{m=1}^M \sum_{n\ne m}\ell_{lc}^{(mn)} + \sum_{m=1}^M \sum_{j=1}^K s_j^m \log s_j^m,
\end{equation}
where $s_j^m = \frac{1}{N}\sum_{i=1}^N q_{ij}^m$.
The first part of Eq.~(\ref{flc}) aims to learn the clustering consistency for all views.
The second part of Eq.~(\ref{flc}) is a regularization term \cite{van2020scan}, which is usually used to avoid all samples being assigned into a single cluster.

Overall, the loss of our multi-view contrastive learning consists of three parts:
\begin{small}
\begin{equation}\label{lc}
\begin{aligned}
    \mathcal{L}&=\mathcal{L}_{\mathbf{Z}} + \mathcal{L}_{\mathbf{H}}+\mathcal{L}_{\mathbf{Q}}  \\
               &=\mathcal{L}_{\mathbf{Z}}(\{\mathbf{X}^m,\mathbf{\hat{X}}^m\}_{m=1}^M; \{\theta^m, \phi^m\}_{m=1}^M)   \\
               &+ \mathcal{L}_{\mathbf{H}}(\{\mathbf{H}^m\}_{m=1}^M;\mathbf{W}_H, \{\theta^m\}_{m=1}^M)\\
               &+ \mathcal{L}_{\mathbf{Q}}(\{\mathbf{Q}^m\}_{m=1}^M;\mathbf{W}_Q, \{\theta^m\}_{m=1}^M),
\end{aligned}
\end{equation}
\end{small}where $\mathcal{L}_{\mathbf{Z}}$ is the reconstruction objective conducted on the low-level features $\{\mathbf{Z}^m\}_{m=1}^M$ to avoid the model collapse.
The consistency objectives $\mathcal{L}_{\mathbf{H}}$ and $\mathcal{L}_{\mathbf{Q}}$ are designed to learn the high-level features and the cluster assignments, respectively.
We learn $\{\mathbf{Q}^m\}_{m=1}^M$ from $\{\mathbf{Z}^m\}_{m=1}^M$ rather than from $\{\mathbf{H}^m\}_{m=1}^M$ as it can avoid the influence between $\mathbf{W}_H$ and $\mathbf{W}_Q$. 
Meanwhile, $\mathbf{W}_H$ and $\mathbf{W}_Q$ will not be influenced by the gradient of $\mathcal{L}_{\mathbf{Z}}$. 
Thanks to this multi-level feature learning structure, we do not need weight parameters to balance the different losses in Eq.~(\ref{lc}) (verified in \cref{sec:Understand}).


\subsection{Semantic Clustering with High-level Features}
Through the multi-view contrastive learning, the model simultaneously learns the high-level features $\{\mathbf{H}^m\}_{m=1}^M$ and the consistent cluster assignments $\{\mathbf{Q}^m\}_{m=1}^M$.
We then treat $\{\mathbf{Q}^m\}_{m=1}^M$ as anchors and match them with the clusters among $\{\mathbf{H}^m\}_{m=1}^M$.
In this way, we can leverage the cluster information contained in the high-level features to improve the clustering effectiveness of the semantic labels.

Concretely, we adopt $K$-means~\cite{macqueen1967some} to obtain the cluster information of each view.
For the $m$-th view, letting $\{\mathbf{c}_k^m\}_{k=1}^K \in \mathbb{R}^{H}$ denote the $K$ cluster centroids, we have:
\begin{equation}\label{kmeans}
    \small \mathop{\min_{\mathbf{c}_1^m,\mathbf{c}_2^m,\dots,\mathbf{c}_K^m}}{\sum_{i=1}^N\sum_{j=1}^K \left\| {\mathbf{h}_i^m-\mathbf{c}_j^m} \right\|_2^2}.
\end{equation}
The cluster labels of all samples $\mathbf p^m \in \mathbb{R}^N$ are obtained by:
\begin{equation}\label{pseudo}
    \small p_i^m = \mathop{\rm{argmin}}_j {\left\| {\mathbf{h}_i^m-\mathbf{c}_j^m} \right\|_2^2}.
\end{equation}

Let $\mathbf{l}^m \in \mathbb{R}^{N}$ denote the cluster labels outputted by the label MLP, where ${l}_i^m=\mathop{\rm{argmax}}_j q_{ij}^m$,
it is worth noting that the clusters represented by $\mathbf p^m$ and $\mathbf{l}^m$ are not corresponding to each other.
Because the clustering consistency is achieved by Eq. (\ref{flc}) in advance, ${l}_i^m$ and ${l}_i^n$ represent the same cluster.
Therefore, we can treat $\mathbf{l}^m$ as anchors to modify $\mathbf p^m$ by the following maximum matching formula:
\begin{small}
\begin{equation}\label{assign}
\begin{aligned}
    &\mathop{\min_{{\mathbf A}^m}}{{\mathbf M^m} {\mathbf A^m}},\\
    s.t.~ &\sum_{i=1} {a}_{ij}^m=1,\sum_{j=1} {a}_{ij}^m=1,\\ 
    &a_{ij}^m\in \{0,1\}, i,j=1,2,...,K,
\end{aligned}
\end{equation}
\end{small}where ${\mathbf A}^m \in \{0,1\}^{K\times K}$ is the boolean matrix and ${\mathbf M}^m \in \mathbb{R}^{K\times K}$ denotes the cost matrix. ${\mathbf M}^m=\mathop{\max_{i,j}}{\tilde{m}}_{ij}^m-{\tilde{\mathbf {M}}}^m$ and ${\tilde{m}}_{ij}^m=\sum_{n=1}^N \mathbbm{1}[{{l}}_n^m=i]\mathbbm{1}[p_n^m=j]$, where $\mathbbm{1}[\cdot]$ represents the indicator function.
Eq.~(\ref{assign}) can be optimized by the Hungarian algorithm~\cite{jonker1986improving}. The modified cluster assignments $\Hat{\mathbf p}_i^m \in \{0,1\}^{K} $ for the $i$-th sample is defined as a one-hot vector. The $k$-th element of $\Hat{\mathbf p}_i^m $ is 1 when $k$ satisfies $k=k\mathbbm{1}[{a}_{ks}^m=1]\mathbbm{1}[p_i^m=s], k,s \in \{1,2,\dots,K\}$.
We then fine-tune the model by cross-entropy loss:
\begin{equation}\label{cross_entropy}
    \small \mathcal{L}_{\mathbf{P}}=-\sum_{m=1}^M \Hat{\mathbf{P}}^m \log \mathbf Q^m,
\end{equation}
where $\Hat{\mathbf{P}}^m=[\Hat{\mathbf p}_1^m;\Hat{\mathbf p}_2^m;\dots;\Hat{\mathbf p}_N^m]\in \mathbb{R}^{N\times K}$.
In this way, we can transfer the learned semantic knowledge to improve the clustering. Finally, the semantic label of the $i$-th sample is:
\begin{equation}\label{y}
    \small y_i = \mathop{\rm{argmax}}_j \left({\frac{1}{M}\sum_{m=1}^M q_{ij}^m}\right).
\end{equation}

\begin{algorithm}[!t]
\caption{: The optimization of MFLVC} 
\label{alg:alg1}
\begin{algorithmic}[1]
\REQUIRE
Multi-view dataset $\{\mathbf{X}^m\}_{m=1}^M$; Number of clusters $K$; Temperature parameters $\tau_F$ and $\tau_L$.

\STATE  Initialize $\{\theta^m,\phi^m\}_{m=1}^M$ by minimizing Eq.~(\ref{lr}).


\STATE  Optimize $\mathbf{W}_H$, $\mathbf{W}_Q$, $\{\theta^m,\phi^m\}_{m=1}^M$ by Eq.~(\ref{lc}).

\STATE  Compute cluster labels by Eqs.~(\ref{kmeans}) and (\ref{pseudo}).
\STATE  Match  multi-view cluster labels by solving Eq.~(\ref{assign}).

\STATE  Fine-tune $\mathbf{W}_Q$, $\{\theta^m\}_{m=1}^M$ by minimizing Eq.~(\ref{cross_entropy}).

\STATE  Calculate semantic labels by Eq.~(\ref{y}).

\ENSURE
The label predictor $\{\{\theta^m\}_{m=1}^M, \mathbf{W}_Q\}$; \\
The high-level feature extractor $\{\{\theta^m\}_{m=1}^M, \mathbf{W}_H\}$.
\end{algorithmic}
\end{algorithm}

\textbf{Optimization}.
The full optimization process of MFLVC is summarized in Algorithm \ref{alg:alg1}.
To be specific, we adopt the algorithm of mini-batch gradient descent to train the model, which consists of multiple autoencoders, a feature MLP, and a label MLP.
The autoencoders are initialized by Eq. (\ref{lr}).
The multi-view contrastive learning is then conducted to achieve the common semantics and clustering consistency by Eq. (\ref{lc}).
After performing the multi-view contrastive learning, the cluster labels obtained from high-level features are modified through the maximum matching formula in Eq.~(\ref{assign}).
The modified cluster labels are then used to fine-tune the model by Eq. (\ref{cross_entropy}).
The high-level feature extractor includes the encoders and the feature MLP, while the label predictor includes the encoders and the label MLP.


\section{Experiments}\label{sec:Experiments}

\subsection{Experimental Setup}

\begin{table}[!h]
\small
\centering
\renewcommand\tabcolsep{10.0pt} 
\begin{threeparttable}
    \begin{tabular}{|c|c|c|c|c|}
    \toprule
    Datasets & \#Samples & \#Views & \#Classes\cr
    \hline
    MNIST-USPS  & 5,000 & 2 & 10 \\
    \hline
    BDGP       & 2,500 & 2 & 5 \\
    \hline
    CCV  & 6,773 & 3 & 20 \\
    \hline
    Fashion & 10,000 & 3 & 10 \\
    \hline
    Caltech-2V & 1,400 & 2 & 7 \\
    \hline
    Caltech-3V & 1,400 & 3 & 7 \\
    \hline
    Caltech-4V & 1,400 & 4 & 7 \\
    \hline
    Caltech-5V & 1,400 & 5 & 7 \\
    \bottomrule
    \end{tabular}
\end{threeparttable}
\caption{The information of the datasets in our experiments.}\label{data}
\end{table}

\begin{table*}[!t]
\small
\centering
\renewcommand\tabcolsep{5.0pt} 
\begin{threeparttable}
    \begin{tabular}{r|ccc|ccc|ccc|ccc}
    \toprule
    Datasets &\multicolumn{3}{c|}{MNIST-USPS} &\multicolumn{3}{c|}{BDGP} &\multicolumn{3}{c|}{CCV} &\multicolumn{3}{c}{Fashion}\cr
    \hline
    Evaluation metrics & ACC & NMI & PUR & ACC & NMI & PUR & ACC & NMI & PUR & ACC & NMI & PUR\\
    \hline
    RMSL \cite{li2019reciprocal} (2019)    & 0.424 & 0.318 & 0.428 & 0.849  & 0.630 & 0.849 & 0.215 & 0.157  & 0.243 & 0.408 & 0.405 & 0.421 \\
    MVC-LFA \cite{wang2019multi} (2019) & 0.768 & 0.675 & 0.768 & 0.564  & 0.395 & 0.612 & 0.232 & 0.195  & 0.261 & 0.791 & 0.759 & 0.794 \\
    COMIC \cite{peng2019comic} (2019)   & 0.482 & 0.709 & 0.531 & 0.578  & 0.642 & 0.639 & 0.157 & 0.081 & 0.157 & 0.578 & 0.642 & 0.608 \\
    CDIMC-net \cite{wen2020cdimc} (2020) & 0.620 & 0.676 & 0.647 & 0.884 & 0.799 & 0.885 & 0.201 & 0.171 & 0.218 & 0.776 & 0.809 & 0.789 \\
    EAMC \cite{zhou2020end} (2020)    & 0.735 & 0.837 & 0.778 & 0.681 & 0.480 & 0.697 & 0.263 & 0.267 & 0.274 & 0.614 & 0.608 & 0.638\\
    IMVTSC-MVI \cite{wen2021unified} (2021)  & 0.669 & 0.592 & 0.717 & \underline{0.981} & \underline{0.950} & \underline{0.982} & 0.117 & 0.060 & 0.158 & 0.632 & 0.648 & 0.635 \\
    SiMVC \cite{trostenMVC} (2021)   & 0.981 & 0.962 & 0.981 & 0.704 & 0.545 & 0.723 & 0.151 & 0.125 & 0.216 & 0.825 & 0.839 & 0.825\\
    CoMVC \cite{trostenMVC} (2021)   & \underline{0.987} & \underline{0.976} & \underline{0.989} & 0.802 & 0.670 & 0.803 & \underline{0.296} & \underline{0.286} & \underline{0.297} & \underline{0.857} & \underline{0.864} & \underline{0.863}\\
    MFLVC (ours)    & \textbf{\textcolor{black}{0.995}} & \textbf{\textcolor{black}{0.985}} & \textbf{\textcolor{black}{0.995}} & \textbf{\textcolor{black}{0.989}}  & \textbf{\textcolor{black}{0.966}} & \textbf{\textcolor{black}{0.989}} & \textbf{\textcolor{black}{0.312}} & \textbf{\textcolor{black}{0.316}}  & \textbf{\textcolor{black}{0.339}} & \textbf{\textcolor{black}{0.992}} & \textbf{\textcolor{black}{0.980}} & \textbf{\textcolor{black}{0.992}} \\
    \bottomrule
    \end{tabular}
\end{threeparttable}
\caption{Results of all methods on four datasets. Bold denotes the best results and underline denotes the second-best.}\label{table1}
\end{table*}

\begin{table*}[!t]
\small
\centering
\renewcommand\tabcolsep{5.0pt} 
\begin{threeparttable}
    \begin{tabular}{r|ccc|ccc|ccc|ccc}
    \toprule
    Datasets &\multicolumn{3}{c|}{Caltech-2V} &\multicolumn{3}{c|}{Caltech-3V} &\multicolumn{3}{c|}{Caltech-4V} &\multicolumn{3}{c}{Caltech-5V}\cr
    \hline
    Evaluation metrics & ACC & NMI & PUR & ACC & NMI & PUR & ACC & NMI & PUR & ACC & NMI & PUR\\
    \hline
    RMSL \cite{li2019reciprocal} (2019) 
    & \underline{0.525}  & 0.474 & 0.540 
    & 0.554  & 0.480 & 0.554
    & 0.596  & 0.551 & 0.608
    & 0.354  & 0.340 & 0.391 \\
    MVC-LFA \cite{wang2019multi} (2019)
    & 0.462  & 0.348  & 0.496 
    & 0.551  & 0.423  & 0.578
    & 0.609  & 0.522  & 0.636 
    & 0.741  & 0.601  & 0.747 \\
    COMIC \cite{peng2019comic} (2019)   
    & 0.422  & 0.446 & 0.535 
    & 0.447 & 0.491 & 0.575 
    & 0.637 & 0.609  & \textbf{\textcolor{black}{0.764}} 
    & 0.532 & 0.549 & 0.604 \\
    CDIMC-net \cite{wen2020cdimc} (2020)   
    & 0.515 & \underline{0.480} & \underline{0.564}
    & 0.528 & 0.483 & 0.565
    & 0.560 & 0.564 & 0.617
    & 0.727 & \underline{0.692} & 0.742 \\
    EAMC \cite{zhou2020end} (2020)   
    & 0.419 & 0.256 & 0.427
    & 0.389 & 0.214 & 0.398
    & 0.356 & 0.205 & 0.370
    & 0.318 & 0.173 & 0.342 \\
    IMVTSC-MVI \cite{wen2021unified} (2021)   
    & 0.490 & 0.398 & 0.540
    & 0.558 & 0.445 & 0.576
    & \underline{0.687} & \underline{0.610} & 0.719
    & \underline{0.760} & 0.691 & \underline{0.785} \\
    SiMVC \cite{trostenMVC} (2021)   
    & 0.508 & 0.471 & 0.557
    & \underline{0.569} & 0.495 & \underline{0.591}
    & 0.619 & 0.536 & 0.630
    & 0.719 & 0.677 & 0.729 \\
    CoMVC \cite{trostenMVC} (2021) 
    & 0.466 & 0.426 & 0.527
    & 0.541 & \underline{0.504} & 0.584
    & 0.568 & 0.569 & 0.646
    & 0.700 & 0.687 & 0.746 \\
    MFLVC (ours)
    & \textbf{\textcolor{black}{0.606}}  & \textbf{\textcolor{black}{0.528}} & \textbf{\textcolor{black}{0.616}} 
    & \textbf{\textcolor{black}{0.631}} & \textbf{\textcolor{black}{0.566}} & \textbf{\textcolor{black}{0.639}} 
    & \textbf{\textcolor{black}{0.733}} & \textbf{\textcolor{black}{0.652}}  & \underline{0.734} 
    & \textbf{\textcolor{black}{0.804}} & \textbf{\textcolor{black}{0.703}} & \textbf{\textcolor{black}{0.804}} \\
    \bottomrule
    \end{tabular}
\end{threeparttable}
\caption{Results of all methods on Caltech with different views. ``-$X$V'' represents that there are $X$ views.}\label{table2}
\end{table*}


\textbf{Datasets}. The experiments are carried out on the five public datasets as shown in Table \ref{data}.
\emph{MNIST-USPS} \cite{peng2019comic} is a popular handwritten digit dataset, which contains 5,000 samples with two different styles of digital images.
\emph{BDGP} \cite{cai2012joint} contains 2,500 samples of drosophila embryos, each of which is represented by visual and textual features.
\emph{Columbia Consumer Video (CCV)} \cite{jiang2011consumer} is a video dataset with 6,773 samples belonging to 20 classes and provides hand-crafted Bag-of-Words representations of three views, such as STIP, SIFT, and MFCC.
\emph{Fashion} \cite{xiao2017fashion} is an image dataset about products, where we follow the literature \cite{Xu_2021_ICCV} to treat different three styles as three views of one product.
\emph{Caltech} \cite{fei2004learning} is a RGB image dataset with multiple views, based on which we build four datasets for evaluating the robustness of the comparison methods in terms of the number of views.
Concretely,
\emph{Caltech-2V} includes WM and CENTRIST;
\emph{Caltech-3V} includes WM, CENTRIST, and LBP;
\emph{Caltech-4V} includes WM, CENTRIST, LBP, and GIST;
\emph{Caltech-5V} includes WM, CENTRIST, LBP, GIST, and HOG.

\textbf{Implementation}.
All the datasets are reshaped into vectors,
and the fully connected networks with the similar architecture are adopted to implement the autoencoders for all views in our MFLVC.
Adam optimizer \cite{kingma2014adam} is adopted for optimization.
The code of MFLVC is implemented by PyTorch \cite{paszke2019pytorch}.
More implementation details are provided in \url{https://github.com/SubmissionsIn/MFLVC}.

\textbf{Comparison methods}.
The comparison methods include classical and state-of-the-art methods, \ie, 4 traditional methods
(RMSL \cite{li2019reciprocal}, MVC-LFA \cite{wang2019multi}, COMIC \cite{peng2019comic}, and IMVTSC \cite{wen2021unified}) and 4 deep methods (CDIMC-net \cite{wen2020cdimc}, EAMC \cite{zhou2020end}, SiMVC \cite{trostenMVC}, and CoMVC \cite{trostenMVC}).

\textbf{Evaluation metrics}. The clustering effectiveness is evaluated by three metrics, \ie, clustering accuracy (ACC), normalized mutual information (NMI), and purity (PUR). 
The mean values of 10 runs are reported for all methods.

\begin{figure*}[!t]
\centering
  \begin{subfigure}{0.24\linewidth}
    \includegraphics[width=1.6in]{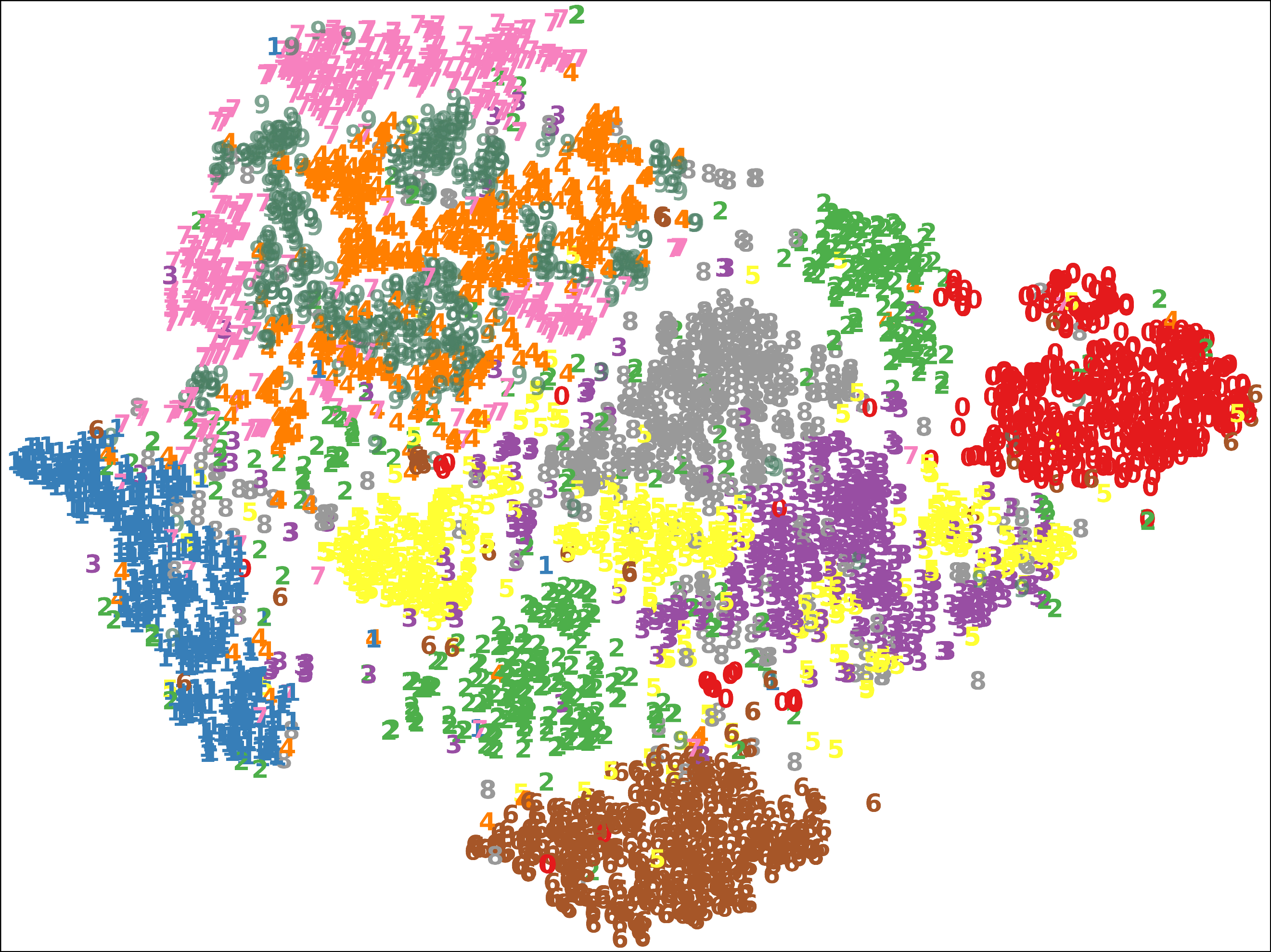}
    \caption{Epoch 0}
  \end{subfigure}
  \hfill
  \begin{subfigure}{0.24\linewidth}
    \includegraphics[width=1.6in]{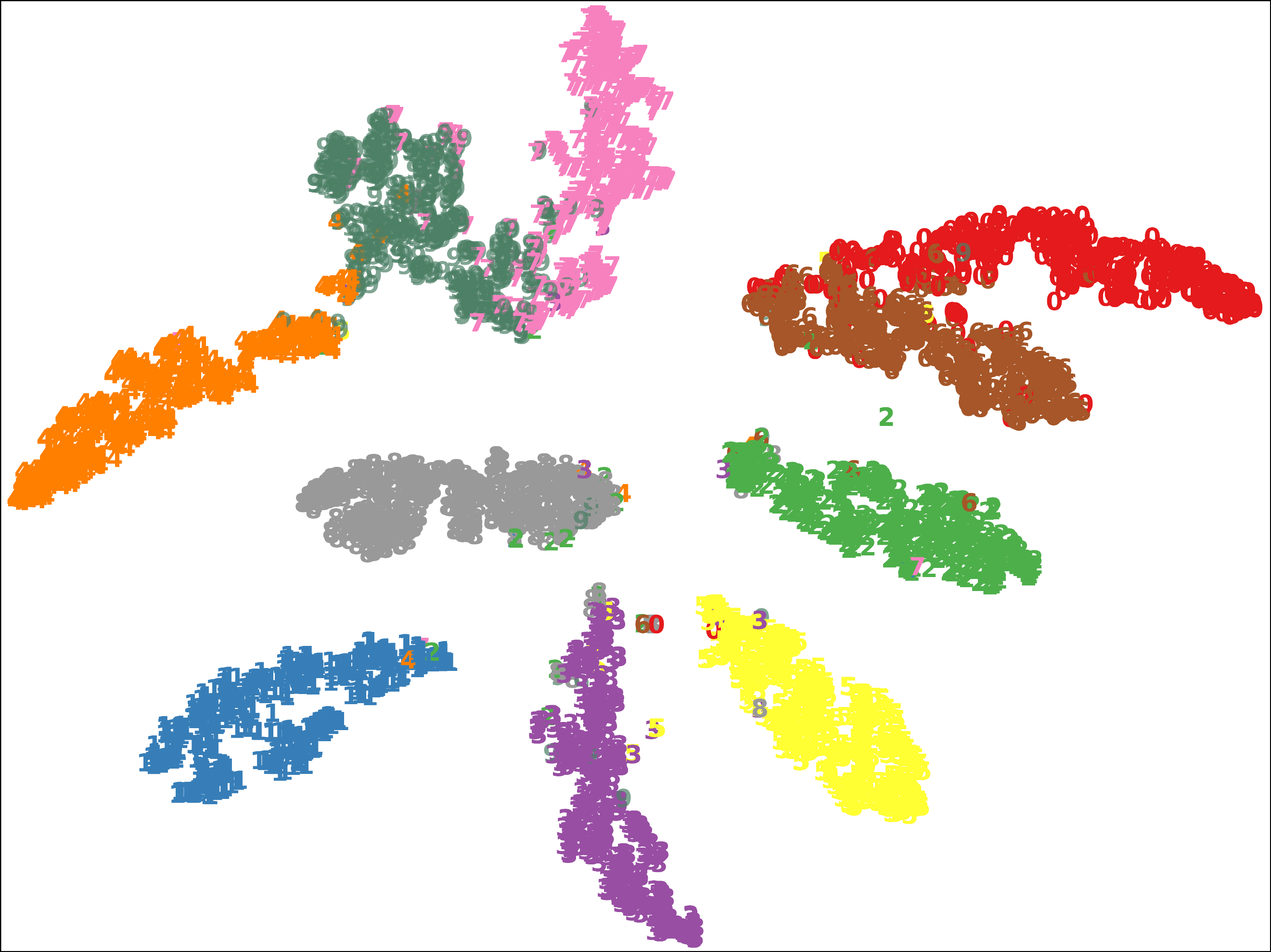}
    \caption{Epoch 5}
  \end{subfigure}
  \hfill
  \begin{subfigure}{0.24\linewidth}
    \includegraphics[width=1.6in]{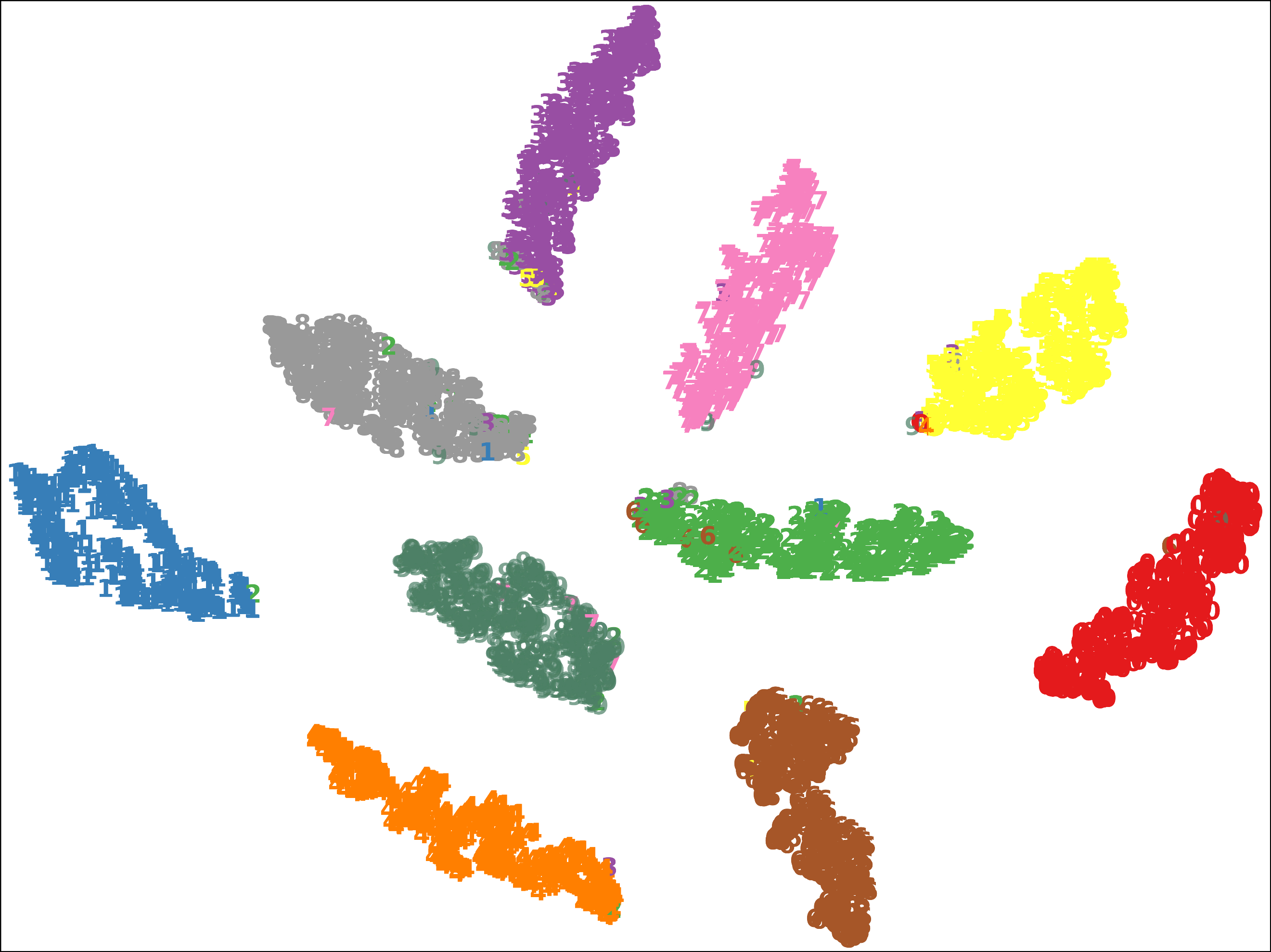}
    \caption{Epoch 10}
  \end{subfigure}
  \hfill
  \begin{subfigure}{0.24\linewidth}
    \includegraphics[width=1.6in]{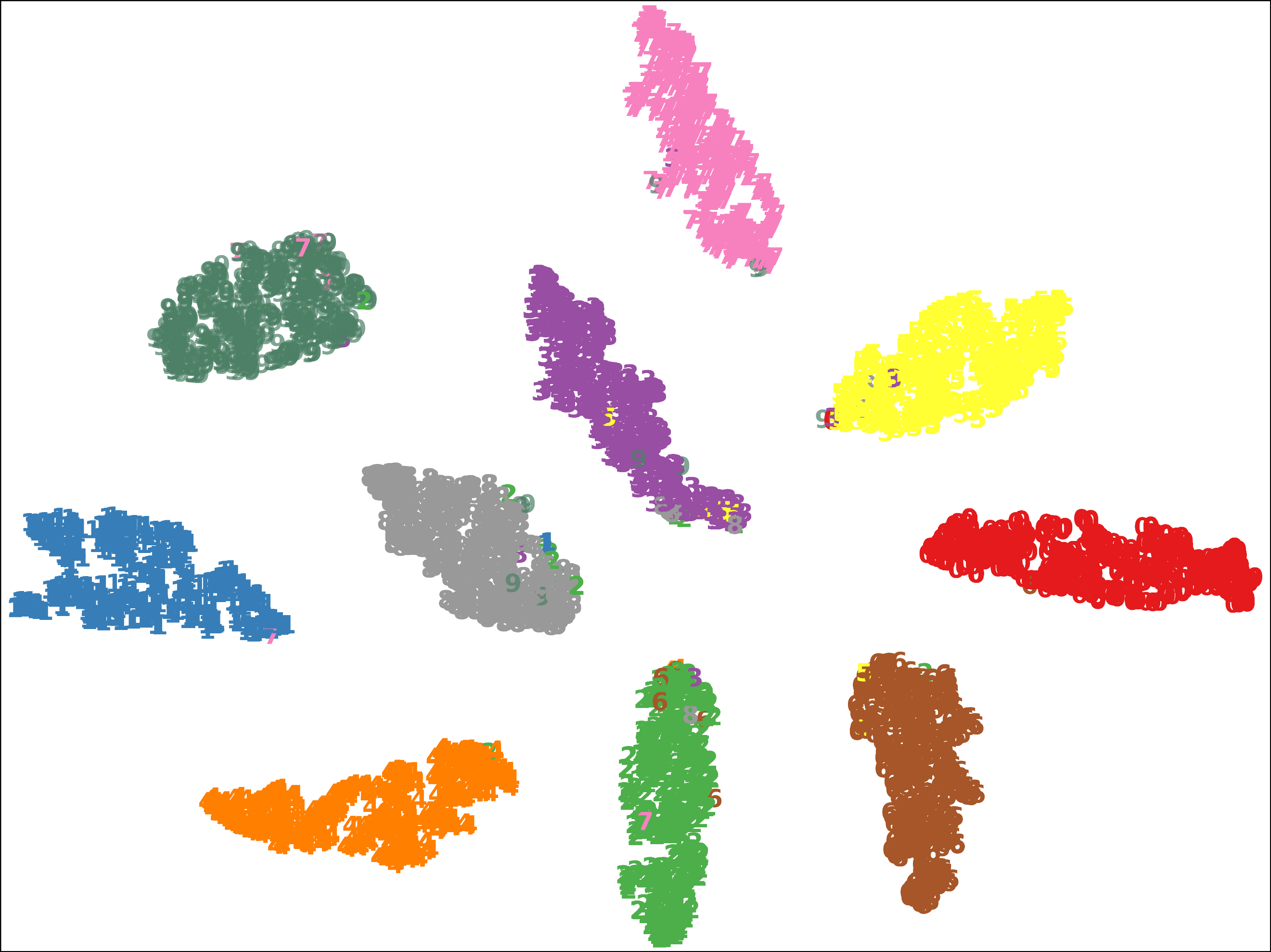}
    \caption{Epoch 15}
  \end{subfigure}
  \vfill
  \begin{subfigure}{0.24\linewidth}
    \includegraphics[width=1.6in]{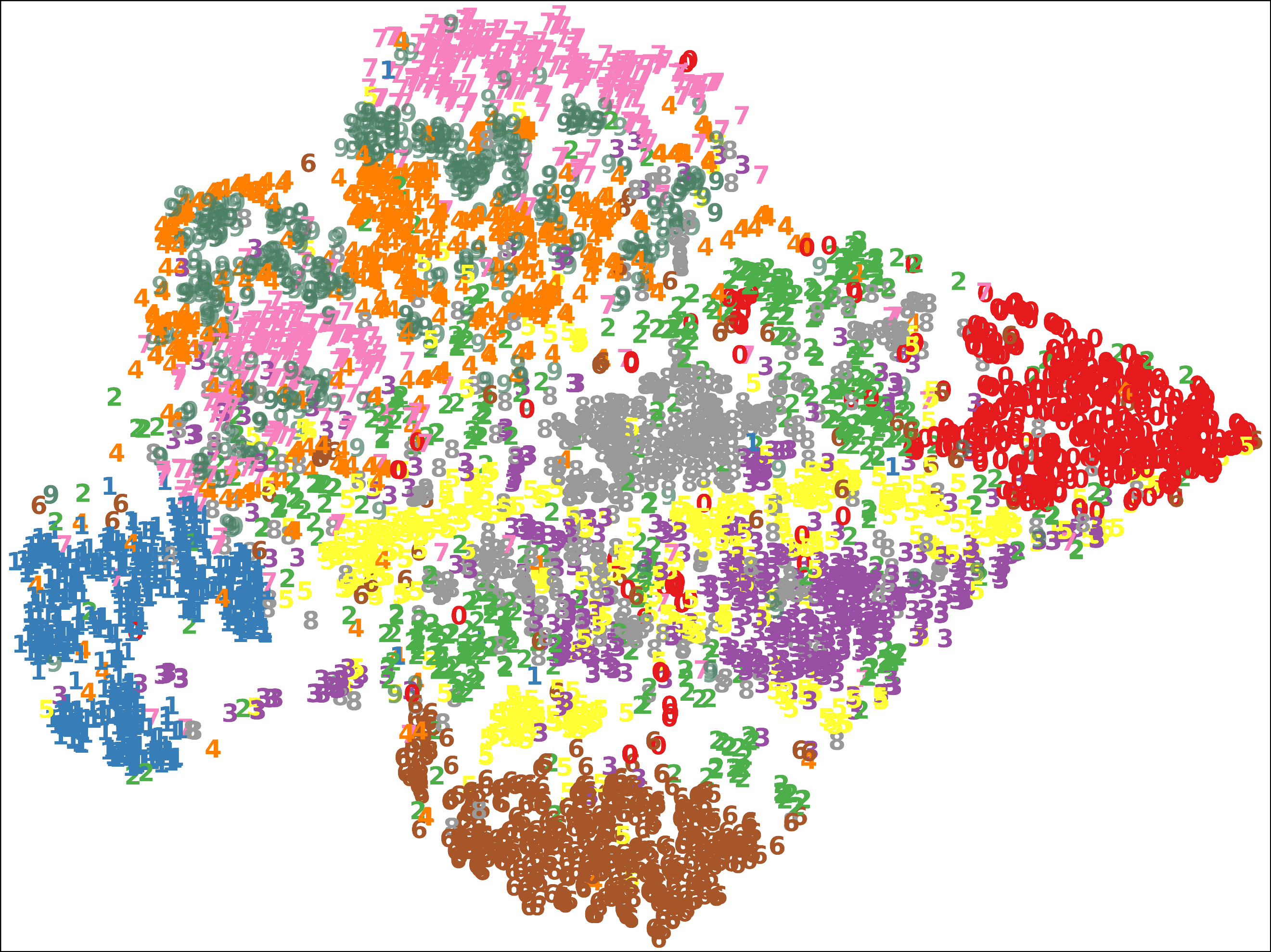}
    \caption{Epoch 0}
  \end{subfigure}
  \hfill
  \begin{subfigure}{0.24\linewidth}
    \includegraphics[width=1.6in]{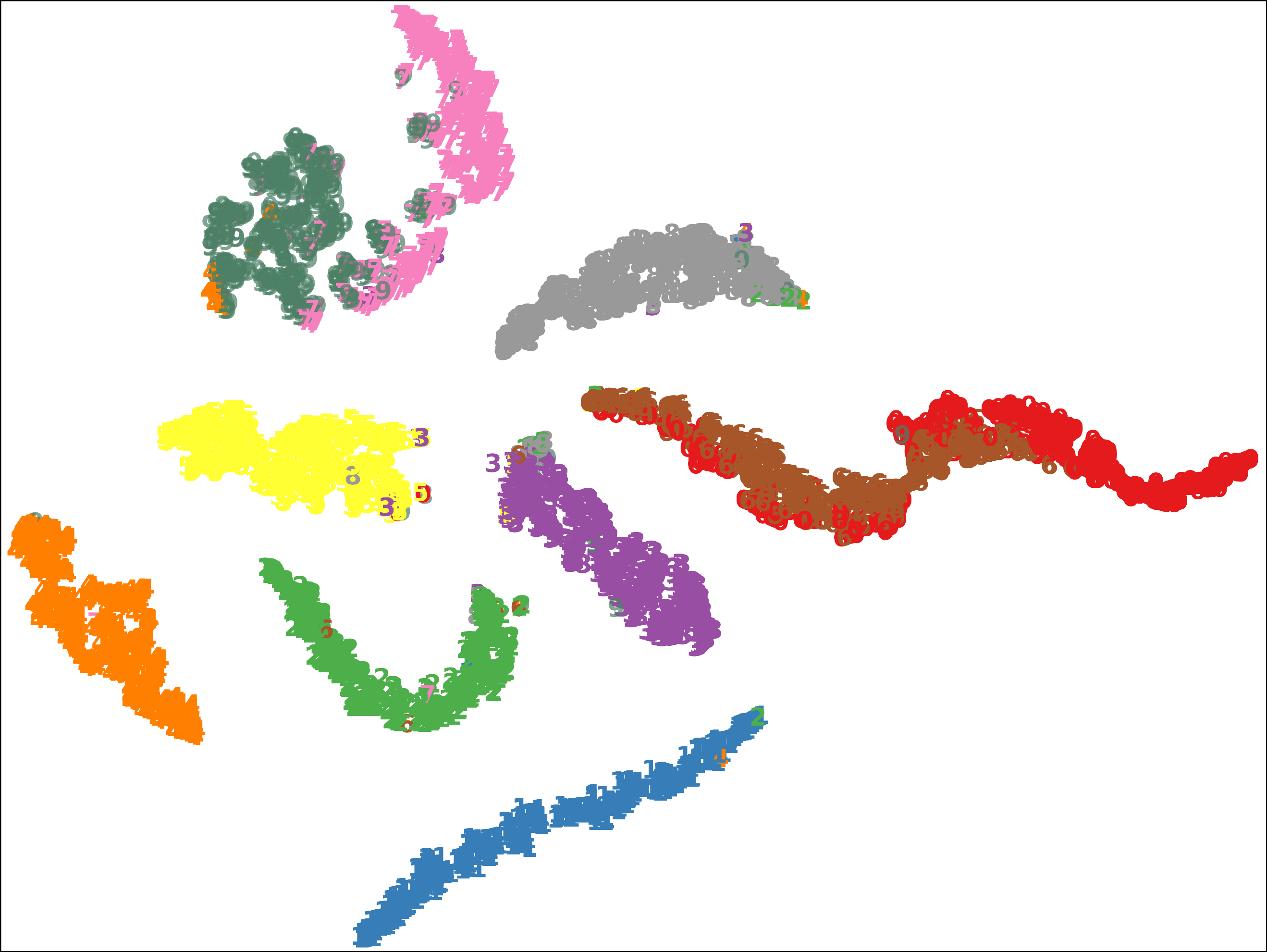}
    \caption{Epoch 5}
  \end{subfigure}
  \hfill
  \begin{subfigure}{0.24\linewidth}
    \includegraphics[width=1.6in]{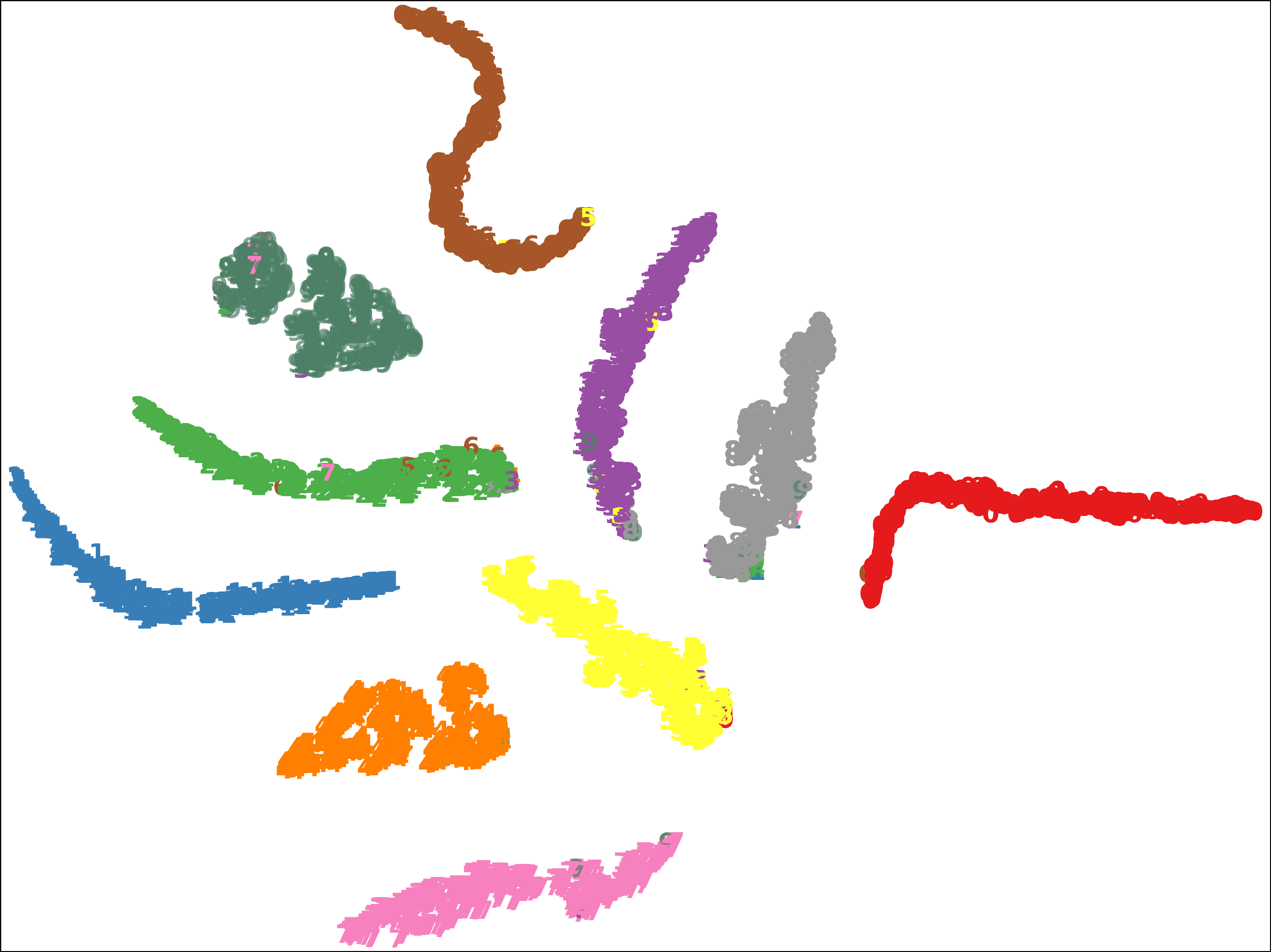}
    \caption{Epoch 10}
  \end{subfigure}
  \hfill
  \begin{subfigure}{0.24\linewidth}
    \includegraphics[width=1.6in]{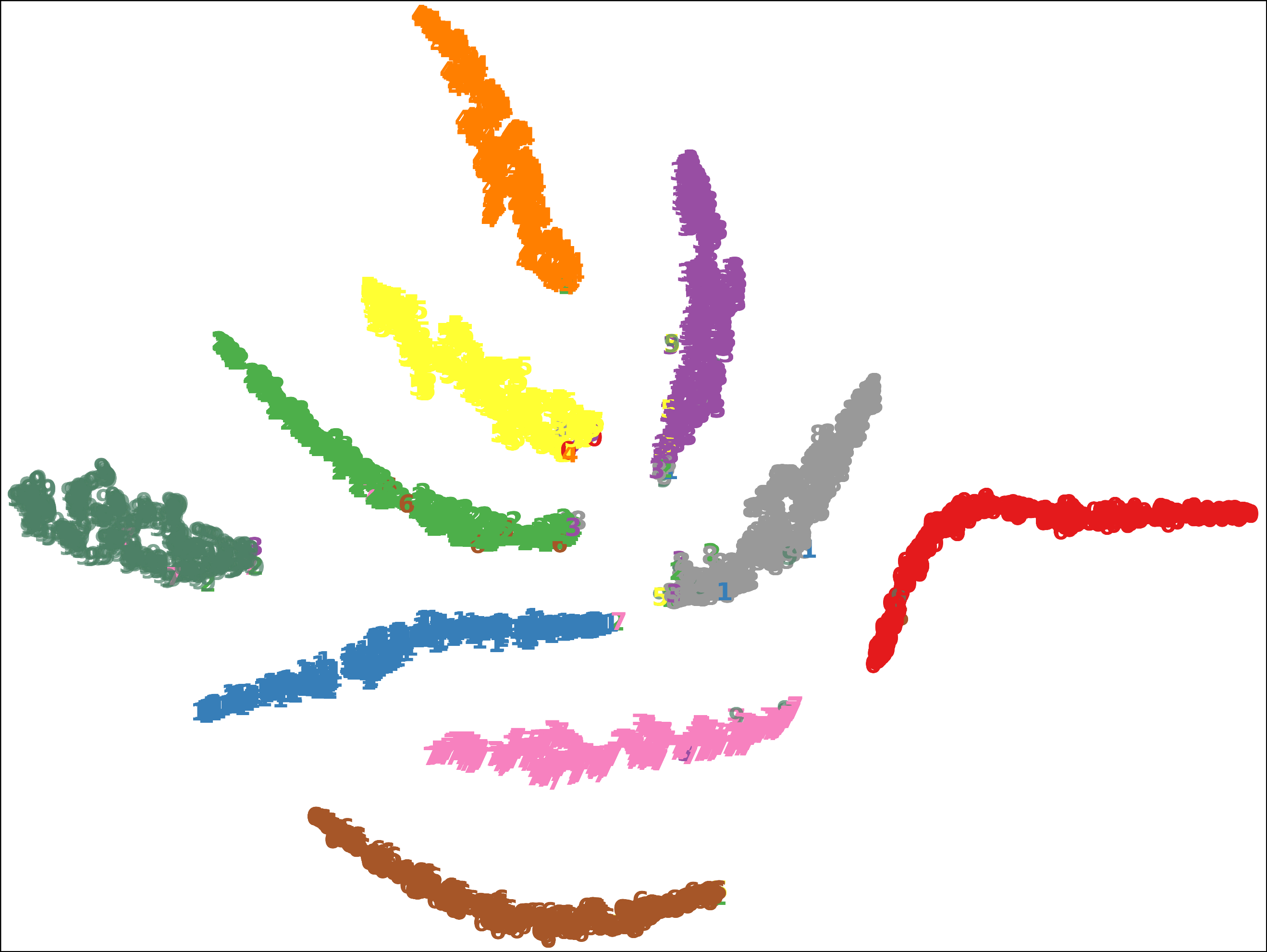}
    \caption{Epoch 15}
  \end{subfigure}
  \caption{Visualization of low-level features (a-d) and high-level features (e-h) for the contrastive learning process.}
  \label{separation}
\end{figure*}

\begin{figure*}[!t]
\centering
  \begin{subfigure}{0.27\linewidth}
    \includegraphics[height=1.15in]{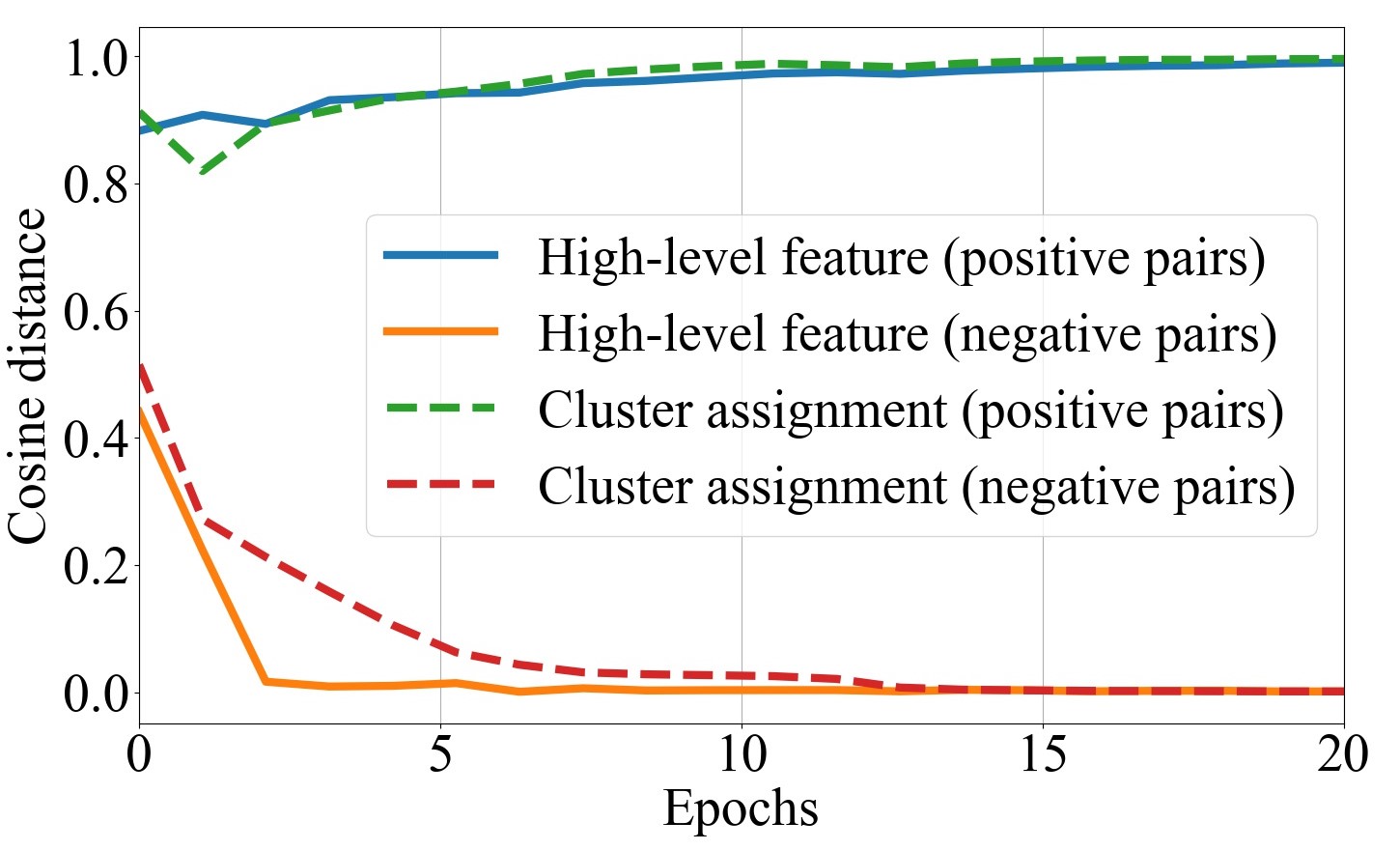}
    \caption{Similarities $vs.$ contrastive learning}
  \end{subfigure}
  \begin{subfigure}{0.25\linewidth}
    \includegraphics[height=1.15in]{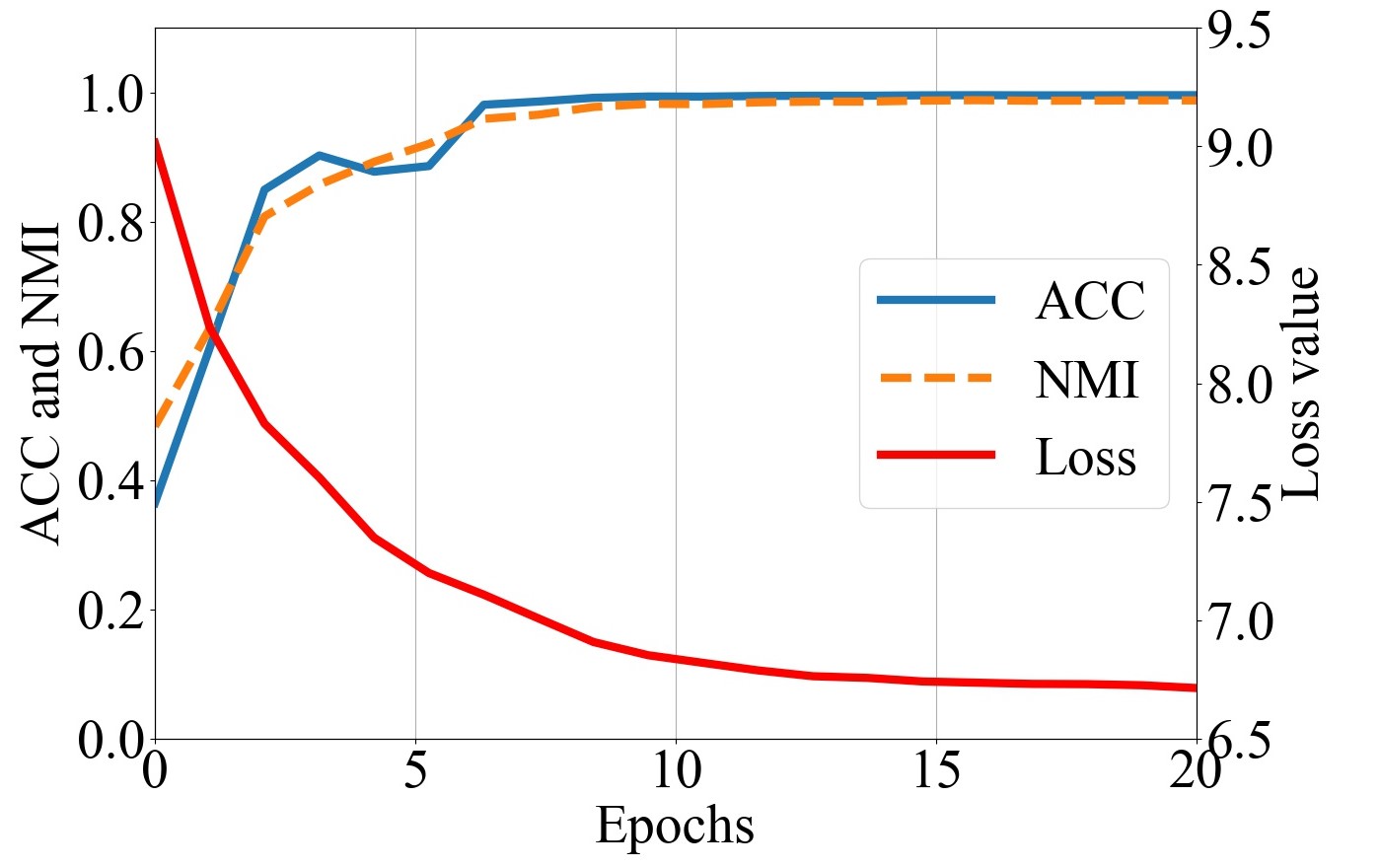}
    \caption{Loss $vs.$ performance}
  \end{subfigure}
  \quad
  \begin{subfigure}{0.20\linewidth}
    \includegraphics[height=1.15in]{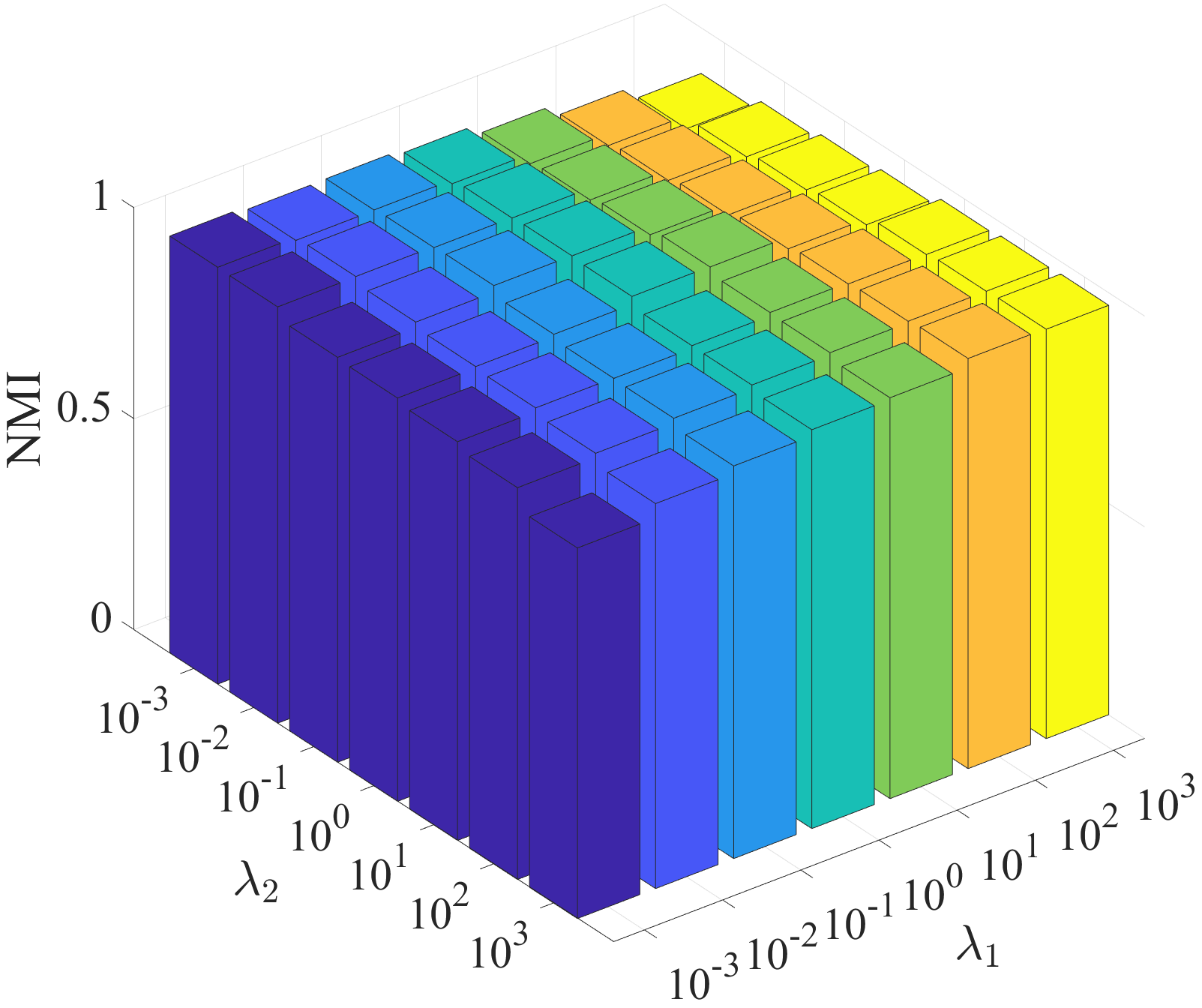}
    \caption{$\lambda_1$ $vs.$ $\lambda_2$}
  \end{subfigure}
  \begin{subfigure}{0.22\linewidth}
    \includegraphics[height=1.15in]{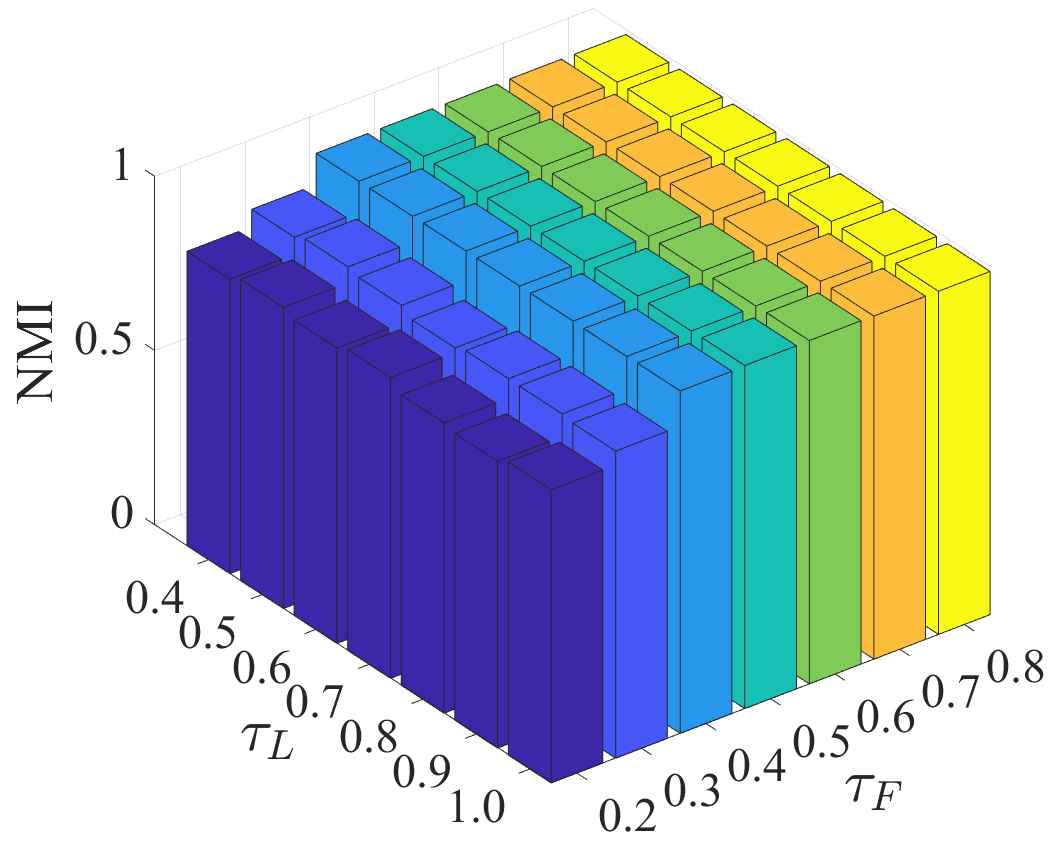}
    \caption{$\tau_F$ $vs.$ $\tau_L$}
  \end{subfigure}
  \caption{(a) The similarities of feature pairs and label pairs. (b) Convergence analysis. (c) and (d) Parameters sensitivity analysis.}\label{visualization}
\end{figure*}

\subsection{Result Analysis}\label{sec:ComparisonsSOTA}
The comparison results on four datasets are shown in Table \ref{table1}, where many comparison methods (\eg, RMSL and COMIC) punish multiple objectives on the same features, and CDIMC-net, EAMC, SiMVC, and CoMVC are feature fusion methods.
One could find that: (1) Our MFLVC achieves the best performance in terms of all metrics. Especially on Dataset Fashion, MFLVC outperforms the best comparison method CoMVC (\ie, 85\%) by about $14\%$ in terms of ACC.
This is because our model is fusion-free and it conducts the reconstruction objective and the consistency objective in different feature spaces so that the adverse influence of view-private information can be reduced.
(2) The improvements obtained by the previous contrastive MVC method (\ie, CoMVC) are limited.
Our MFLVC is also a contrastive MVC method, instead, it avoids the fusion of view-private information and its multi-level feature learning framework allows the high-level features to learn the common semantics across all views more effectively.

To further verify our method, we build four datasets based on Caltech and test the performance of all comparison methods.
Table~\ref{table2} shows the results on Caltech with different views, from which we could have the following observations:
(1) The clustering effectiveness of most methods improves with the increase of the number of views, \ie, ACC increases from $60\%$ to $80\%$.
(2) Compared to 8 comparison methods, our MFLVC mostly achieves the best performance indicating its robustness.
(3) Some methods obtain bad results when increasing the number of views. 
For example, RMSL, COMIC, and EAMC achieve ACC about $35\%$, $53\%$, and $31\%$  on Caltech-5V which are lower than that on Caltech-4V (\ie, $59\%$, $63\%$, and $35\%$).
The reason is that the data of each view simultaneously contain useful common semantics as well as meaningless view-private information.
Views contain much view-private information which might increase the difficulty of extracting their common semantics.
These observations further verify the effectiveness of our method, which learns multiple levels of features so as to reduce the interference from the view-private information.

\section{Model Analysis}



\subsection{Understand the Multi-level Feature Learning}\label{sec:Understand}
In order to investigate the proposed multi-level feature learning, we take MNIST-USPS as an example and visualize its training process.
The MNIST view is shown in Figure~\ref{separation} via $t$-SNE~\cite{maaten2008visualizing}.
It can be discovered that the cluster structures of low-level features and high-level features become clear during the training process. 
The clusters of low-level features are not dense.
This is because the low-level features have maintained the diversity among samples by reconstruction objective.
In contrast, the clusters of high-level features are dense and have better low-dimensional manifolds. 
Additionally, in Figure~\ref{visualization}(a), the similarities of positive feature pairs are rising while that of negative feature pairs are decreasing. 
This indicates that the information learned by the high-level features is close to the common semantics across multiple views.
These observations are in agreement with our motivations, \ie, the feature MLP can filter out the view-private information of multiple views so the outputted high-level features are in dense shapes.
The similarities of positive label pairs are also rising which indicates that the clustering consistency of semantic labels is achieved.




\textbf{Convergence analysis}.
It is not difficult to discover that the objectives of $\mathcal{L}_{\mathbf{Z}}$, $\mathcal{L}_{\mathbf{H}}$, $\mathcal{L}_{\mathbf{Q}}$, and $\mathcal{L}_{\mathbf{P}}$, \ie, Eqs. (\ref{lr},\ref{lfr},\ref{flc},\ref{cross_entropy}) are all convex functions. As shown in Figure~\ref{visualization}(b), the clustering effectiveness increases with the decrease of loss values, indicating that MFLVC enjoys good convergence property.

\textbf{Parameter sensitivity analysis}.
We investigate  whether hyper-parameters are needed to balance the loss components in Eq.~(\ref{lc}), \ie, $\mathcal{L}_{\mathbf{Z}} + \lambda_1\mathcal{L}_{\mathbf{H}}+\lambda_2\mathcal{L}_{\mathbf{Q}}$. Figure~\ref{visualization}(c) shows the mean values of NMI within 10 independent runs, which indicates that our model is insensitive to $\lambda_1$ and $\lambda_2$.
This is because our model has a well-designed multi-level feature learning framework, by which the interference among different features can be reduced.
In this paper, we set $\lambda_1 = 1.0 $ and $\lambda_2 = 1.0$ for all used datasets.
Furthermore, the multi-view contrastive learning includes two temperature parameters, \ie, $\tau_F$ of the feature contrastive loss in Eq.~(\ref{fccc}) and $\tau_L$ of the label contrastive loss in Eq.~(\ref{lccc}).
Figure~\ref{visualization}(d) indicates that our model is insensitive to the choice of $\tau_F$ and $\tau_L$.
Empirically, we set $\tau_F = 0.5$ and $\tau_L = 1.0$.

\subsection{Ablation Studies}\label{sec:Ablation}

\textbf{Loss components}.
We conduct ablation studies on the loss components in Eq. (\ref{lc}) and Eq. (\ref{cross_entropy}) to investigate their effectiveness.
Table~\ref{losscomponents} shows  different loss components and the corresponding experimental results.
(A) $\mathcal{L}_{\mathbf{Q}}$ is optimized to achieve the basic goal of multi-view clustering, \ie, learning the clustering consistency.
(B) $\mathcal{L}_{\mathbf{Z}}$ is optimized to make the low-level features be capable of reconstructing the multiple views.
(C) $\mathcal{L}_{\mathbf{H}}$ is optimized to learn the high-level features, which are then used to fine-tune the semantic labels by $\mathcal{L}_{\mathbf{P}}$.
(D) The complete loss components of our method.
In terms of the results, (B) and (D) have better performance than (A) and (C), respectively, indicating that the reconstruction objective is important.
Especially when the model has only low-level features, the results of (B) are better than that of (A) by about $20\%$ and $10\%$ on MNIST-USPS and BDGP, respectively.
According to (C) and (D), we can find that the learned high-level features play the most important role in improving the clustering effectiveness.
For example, the results of (C) are better than that of (A) by about $30\%$ and $20\%$ on MNIST-USPS and BDGP, respectively.


\begin{table}[!t]
\small
\centering
\renewcommand\tabcolsep{3.9pt}
\begin{threeparttable}
    \begin{tabular}{l|c|c|c|cc|cc}
    \toprule
    &\multicolumn{3}{c|}{Components} &\multicolumn{2}{c|}{MNIST-USPS} &\multicolumn{2}{c}{BDGP}\cr
    \hline
    &$\mathcal{L}_{\mathbf{Q}}$ &$\mathcal{L}_{\mathbf{Z}}$ &$\mathcal{L}_{\mathbf{H}}$ and $\mathcal{L}_{\mathbf{P}}$ & ACC & NMI & ACC & NMI \\
    \hline
    (A) &$\checkmark$ &\quad &\quad     &0.676 &0.777 &0.715 &0.663\\
    \hline
    (B) &$\checkmark$ &$\checkmark$ &\quad   &0.891 &0.939 &0.825 &0.690\\
    \hline
    (C) &$\checkmark$ &\quad &$\checkmark$   &0.984 &0.962 &0.955 &0.886\\
    \hline
    (D) &$\checkmark$ &$\checkmark$&$\checkmark$  &0.995 &0.985 &0.989 &0.966\\
    \bottomrule
    \end{tabular}
\end{threeparttable}
\caption{Ablation studies on loss components.}\label{losscomponents}
\end{table}


\begin{table}[!t]
\small
\centering
\renewcommand\tabcolsep{4.2pt}
\begin{threeparttable}
    \begin{tabular}{l|l|cc|cc}
    \toprule
    & &\multicolumn{2}{c|}{MNIST-USPS} &\multicolumn{2}{c}{BDGP}\cr
    \hline
    & & ACC & NMI & ACC & NMI \\
    \hline
    (a) & $\mathbf{X}-\mathbf{Q}_{\checkmark}$         &0.676 &0.777 &0.715 &0.663 \\
    \hline
    (b) & $\mathbf{X}-\mathbf{Z}_{\checkmark}-\mathbf{Q}_{\checkmark}$   &0.921 &0.860 &0.652 &0.498 \\
    \hline
    (c) & $\mathbf{X}-\mathbf{Z}_{\checkmark}-\mathbf{H}_{\checkmark}-\mathbf{Q}_{\checkmark}$  &0.948 &0.894 &0.742 &0.654 \\
    \hline
    (d) & $\mathbf{X}-\mathbf{Z}_{\times}-\mathbf{H}_{\checkmark}-\mathbf{Q}_{\checkmark}$  &0.995 &0.985 &0.989 &0.966 \\
    \bottomrule
    \end{tabular}
\end{threeparttable}
\caption{Ablation studies on contrastive learning structures. ``$\checkmark$'' represents that the contrastive loss is optimized on the features.}
\label{contrastive}
\end{table}

\textbf{Contrastive learning structures}.
To further verify our proposal, we perform contrastive learning (\ie, consistency objective) on different network structures.
As shown in Table~\ref{contrastive},
(a) The semantic labels $\mathbf{Q}$ are learned directly from the input features $\mathbf{X}$.
This structure is similar to \cite{zhong2020deep,van2020scan,niu2021spice} in some degree.
It results in poor performance by directly extending contrastive learning to the multi-view scenarios.
(b) Between $\mathbf{X}$ and $\mathbf{Q}$, we set the low-level features $\mathbf{Z}$ and perform contrastive learning on $\mathbf{Q}$ and $\mathbf{Z}$.
This structure is similar to \cite{li2021contrastive,lin2021completer,trostenMVC} in some degree and the performance is also limited.
(c) Based on $\mathbf{Z}$, we stack a feature MLP to obtain the high-level features $\mathbf{H}$ and perform contrastive learning on $\mathbf{Z}$, $\mathbf{H}$, and $\mathbf{Q}$.
As for (b) and (c), the reconstruction objective is also performed on $\mathbf{Z}$.
(b) and (c) make progress on MNIST-USPS, because the two views of MNIST-USPS are digital images and they have little view-private information to influence the learning performance.
However, (b) and (c) cannot mine the common semantics well on BDGP. The reason is that the two views of BDGP are visual features and text features and they have much view-private information.
It results in poor performance when performing reconstruction and consistency objectives on the same features (\ie, $\mathbf{Z}$).
(d) We perform contrastive learning only on $\mathbf{H}$ and $\mathbf{Q}$ while leaving reconstruction objective on $\mathbf{Z}$.
This setting obtains the best performance by performing consistency and reconstruction objectives in different feature spaces.
These experiments further verified the effectiveness of our method, and confirmed that it is useful to learn representations via a multi-level feature learning structure.




\section{Conclusion}
In this paper, we have proposed a new framework of multi-level feature learning for contrastive multi-view clustering.
For each view, the proposed framework learns multiple levels of features, including low-level features, high-level features, and semantic labels in a fusion-free manner.
This allows our model to learn the common semantics across all views and reduce the adverse influence of view-private information.
Extensive experiments on five public datasets demonstrate that our method obtains state-of-the-art performance.

\textbf{Broader impacts}. The proposed framework learned a high-level feature extractor and a label predictor, which can be applied to downstream tasks such as feature compression, unsupervised labeling, and cross-modal retrieval, \etc. However, this work aims to provide a general framework and the trained model might be affected by the intrinsic bias of data especially with dirty samples. Therefore, the future works could extend our framework to other application scenarios.


\section*{Acknowledgments}
This work was partially supported by the National Natural Science Foundation of China (Grants No. 61806043 and No. 61876046) and the Guangxi ``Bagui'' Teams for Innovation and Research, China. Lifang He was supported by the Lehigh's Accelerator Grant (No. S00010293).

{\small
\bibliographystyle{ieee_fullname}
\bibliography{MAIN}
}

\end{document}